\documentclass{article}
\usepackage{iclr2027_conference,times}

\usepackage{amsmath,amsfonts,amssymb}
\usepackage{booktabs,multirow,tabularx}
\usepackage{graphicx,subcaption}
\usepackage{xcolor}
\usepackage{float,needspace}
\usepackage{hyperref}
\usepackage{fontawesome5}

\hypersetup{
  hidelinks,
  pdftitle={FireNav: Decision-Oriented Reconstruction Evidence Routing
    for Vision-Language Navigation in Indoor Fire Environments},
  pdfauthor={Hongbo Ding, Xiang Chen, Jiangxin Dong, Jinshan Pan}
}

\usepackage{amsmath,amsfonts,bm}

\def\eqref#1{equation~\ref{#1}}
\def\1{\bm{1}}

\DeclareMathAlphabet{\mathsfit}{\encodingdefault}{\sfdefault}{m}{sl}
\SetMathAlphabet{\mathsfit}{bold}{\encodingdefault}{\sfdefault}{bx}{n}

\title{FireNav: Decision-Oriented Reconstruction Evidence Routing for Vision-Language Navigation in Indoor Fire Environments}

\author{%
\hspace*{-\tabcolsep}%
\parbox[t]{\textwidth}{%
\centering
\normalfont
{\textbf{Hongbo Ding}}\textsuperscript{1}\quad
\href{https://cschenxiang.github.io/}{\textbf{Xiang Chen}}\textsuperscript{2,*}\quad
\href{https://scholar.google.com/citations?user=ruebFVEAAAAJ}{\textbf{Jiangxin Dong}}\textsuperscript{2}\quad
\href{https://scholar.google.com/citations?user=CMsNjGIAAAAJ}{\textbf{Jinshan Pan}}\textsuperscript{2}
\\[5pt]
\normalfont
\textsuperscript{1}\href{https://lowlevelcv.com/}{Nanjing EntroVision Technology Co., Ltd}
\\[2pt]
\textsuperscript{2}Nanjing University of Science and Technology
\\[3pt]
{\normalfont\small\textcolor{gray}{\textsuperscript{*}Corresponding author}}
\\[4pt]
{\small
\href{https://firenav-bench.github.io/FireNav/\#overview}{%
\textcolor{black}{\faHome}\enspace
\textcolor{blue}{\textbf{\texttt{Project Homepage}}}%
}}
}%
\hspace*{-\tabcolsep}%
}

\iclrfinalcopy

\begin{document}
\maketitle

% 必须放在 \maketitle 之后：
% 模板会在生成标题时重新设置页眉。
\pagestyle{fancy}
\fancyhf{}
% 清空页眉，仍保留底部页码：
\fancyhead{}
%\fancyhead[L]{Under review}
\fancyfoot[C]{\thepage}
\renewcommand{\headrulewidth}{0pt}
\renewcommand{\footrulewidth}{0pt}

\vspace{-5mm}

\begin{abstract}
In real-world indoor fire environments, visual degradation severely impairs instruction grounding, waypoint prediction, and long-horizon exploration in vision-language navigation (VLN).
Although image restoration offers a natural means to mitigate such visual degradation, existing methods primarily optimize reconstruction fidelity, which does not necessarily translate into better navigation decisions and may even corrupt task-critical cues when applied indiscriminately.
In this paper, we present FireNav, a decision-oriented reconstruction evidence routing framework that aligns visual restoration with navigation utility in VLN.
To advance this task, we first contribute FireVLN-Bench, a paired indoor fire navigation benchmark with clean references preserving task semantics, safety supervision, and matched settings for continuous navigation and robot execution with simulated dynamics.
Our FireNav retains native visual tokens and treats latent reconstruction features as auxiliary evidence, where a lightweight router leverages visual features, task context, and prior execution states to determine whether such evidence should be admitted and how strongly it should contribute to the visual tokens.
To learn when reconstruction is beneficial, we further introduce decision utility calibration, which compares action prediction losses with and without candidate evidence under the same state, thereby supervising evidence selection according to its contribution to waypoint, heading, and arrival prediction.
Experiments demonstrate that FireNav consistently outperforms state-of-the-art baselines in both navigation success and path efficiency across evaluation settings.
\end{abstract}

\section{Introduction}
\label{sec:introduction}
Vision-language navigation (VLN) requires embodied agents to ground natural language instructions in visual observations and translate them into navigation actions~\citep{janusvln,awarevln}. 
Recent methods have substantially advanced spatial reasoning, state modeling, and action planning, enabling increasingly capable navigation in complex indoor environments~\citep{omninav,dualvln}. 
However, most existing VLN methods assume clear or nominal visual observations, leaving robustness under adverse visual conditions less explored.

Adverse visual conditions in embodied navigation can propagate beyond perception to reasoning, decision making, and action execution~\citep{embodiediqa,alldaywalker}. 
Such errors may induce incorrect waypoints or stopping decisions and further affect subsequent observations~\citep{wang2025vlnpe,s2e}.
This issue is particularly critical in indoor fire environments, where smoke and poor visibility may obscure doorways, people, signs, or obstacle boundaries. 
Therefore, visual reconstruction for VLN should not only improve visibility, but also preserve task-relevant hazards, targets, and structural cues required for reliable navigation.

A natural solution is to directly apply existing image restoration methods to degraded observations before navigation. 
Recent approaches can handle heterogeneous degradations through task-aware prompting, adaptive routing, and degradation priors~\citep{datprlir,unirestorer,unirestore,r2rrestore}. 
However, better reconstruction quality does not necessarily lead to better navigation decisions. 
Restoring every observation may introduce unnecessary computation, enhancing irrelevant regions may not benefit the next action, and unconditional fusion of reconstructed features may even disturb reliable native representations. 
This motivates a key question: \emph{when should reconstruction influence VLN, and how strongly should it modify the native visual representation?}

In this paper, we investigate robust VLN in indoor fire environments. Specifically, we first introduce \textbf{FireVLN-Bench}, a paired benchmark that combines diverse indoor scenes, hazard layouts, and evolving visual degradations with clean references that preserve fire- and task-relevant semantics. 
Note that paired observations share the same camera poses and scene states, while matched episodes support continuous navigation and robot execution under simulated dynamics. 
Our benchmark provides shortest and risk-aware trajectories, enabling evaluation of both navigation efficiency and safety under consistent execution settings.

Building on FireVLN-Bench, we present \textbf{FireNav}, a robust framework for decision-oriented reconstruction evidence routing. 
Our method retains native visual tokens and treats latent reconstruction features as auxiliary evidence. Cross-attention extracts reconstruction evidence within each view, while a lightweight router uses visual features, task context, and prior execution states to determine where and how strongly such evidence should influence navigation. 
Different from existing methods that supervise decisions through action quality or task progress~\citep{compassnav,progressthink,awarevln,adaptivechunking}, FireNav explicitly measures the incremental utility of reconstruction for the current decision. 
At the same navigation state, predictions with and without candidate reconstruction evidence are compared through waypoint, heading, and arrival losses to supervise evidence routing. 
For efficient evidence integration, we adopt bounded residual fusion with zero-initialized evidence contribution and regularize the policy on clean observations, while incremental feature caching avoids redundant visual encoding.
Experiments show that FireNav improves navigation success and path efficiency over reconstructed-RGB and capacity-matched adapter baselines, while preserving performance on clean observations with limited inference overhead.

Our main contributions are summarized as follows.
\begin{itemize}
\item We construct a high-quality VLN benchmark, FireVLN-Bench, for indoor fire navigation with task-preserving clean references, matched degraded observations, and consistent continuous and physical navigation settings.

\item We propose FireNav, a decision-oriented reconstruction evidence routing framework that selectively injects reconstruction features into a frozen navigator while preserving its native visual interface for robust indoor fire VLN.

\item We develop decision utility calibration and bounded residual fusion to supervise evidence admission based on its impact on action prediction and to constrain the resulting changes to native visual representations.

\item Extensive experiments demonstrate that FireNav consistently improves navigation success and path efficiency under indoor fire conditions, establishing an effective bridge between visual reconstruction and robust VLN.
\end{itemize}

\section{Related Work}
\label{sec:related-work}

\paragraph{Vision-Language Navigation.}
VLN maps visual observations conditioned on language to navigation actions and now spans discrete graph navigation, continuous control, open vocabulary search, persistent exploration, and vision-language-action policies. Recently, OmniNav~\citep{omninav} unifies instruction goal, object goal, and point goal navigation with frontier exploration through a fast policy for continuous waypoints and a slow planner with memory. DualVLN~\citep{dualvln} adopts a dual system design that decouples VLM grounding from frequent trajectory execution using explicit pixel goals and implicit latent goals. 
However, most mainstream VLN methods assume clear or nominal visual observations, leaving robustness under adverse visual conditions less explored. AlldayWalker~\citep{alldaywalker} addresses this problem by adapting a navigator across low light, overexposed, and scattering environments while mitigating forgetting across scenarios. Such adaptation highlights the need for robust visual representations but does not study how reconstruction features can support closed loop navigation without replacing the pretrained visual pathway. In this paper, we explore robust VLN under degradation in indoor fire environments, where compound visual degradations jointly challenge perception and decision making.

\vspace{-2mm}

\paragraph{Visual Reconstruction.}
Visual reconstruction aims to recover visibility, structure, and appearance from degraded observations, with most existing methods optimizing image fidelity or perceptual quality. This objective also extends to fire scenes obscured by smoke. Huang et al.~\citep{desmokegan} propose DesmokeGAN, which uses attention-guided adversarial learning for image desmoking and smoke detection, while SmokeBench~\citep{smokebench} provides aligned real-world smoke/clean surveillance pairs for early-stage fire desmoking. Given the shared scattering and contrast attenuation effects of smoke and haze, recent dehazing methods also offer relevant reconstruction priors. IPC-Dehaze~\citep{ipcdhaze} iteratively refines discrete visual codes for real-world dehazing, while DCL~\citep{dcl} couples atmospheric scattering with temporal brightness consistency via shared depth estimation. Despite improving visibility and perception, these methods largely treat reconstructed images as final outputs. In contrast, embodied VLN requires decision-oriented reconstruction. Thus, we use reconstruction features as optional evidence and evaluate their utility in closed-loop navigation through navigation supervision and task-preservation constraints.

\vspace{-2mm}

\paragraph{Evidence Routing.}
Moving from perceptual restoration to decision-oriented reconstruction raises a further question of which recovered features should influence the task representation. UniRestore~\citep{unirestore} extends restoration toward downstream tasks through task aware feature adaptation and fusion. TaskTok~\citep{tasktok} selectively refines restoration tokens relevant to the task rather than treating the full latent sequence uniformly. This selective use of latent evidence also appears in embodied decision making. DualVLN~\citep{dualvln} transfers compact latent goals from a slow VLM planner to a fast trajectory policy. x$^2$-Fusion~\citep{x2fusion} adjusts heterogeneous feature contributions according to their reliability under degradation. In indoor fire scenes, reconstruction cues may recover obscured evidence but also distort semantics critical to the task. FireNav therefore introduces decision-guided reconstruction evidence routing to use reconstruction evidence only when it is reliable and beneficial to navigation decisions.

\begin{figure*}[t]
    \centering
    \includegraphics[width=\textwidth]{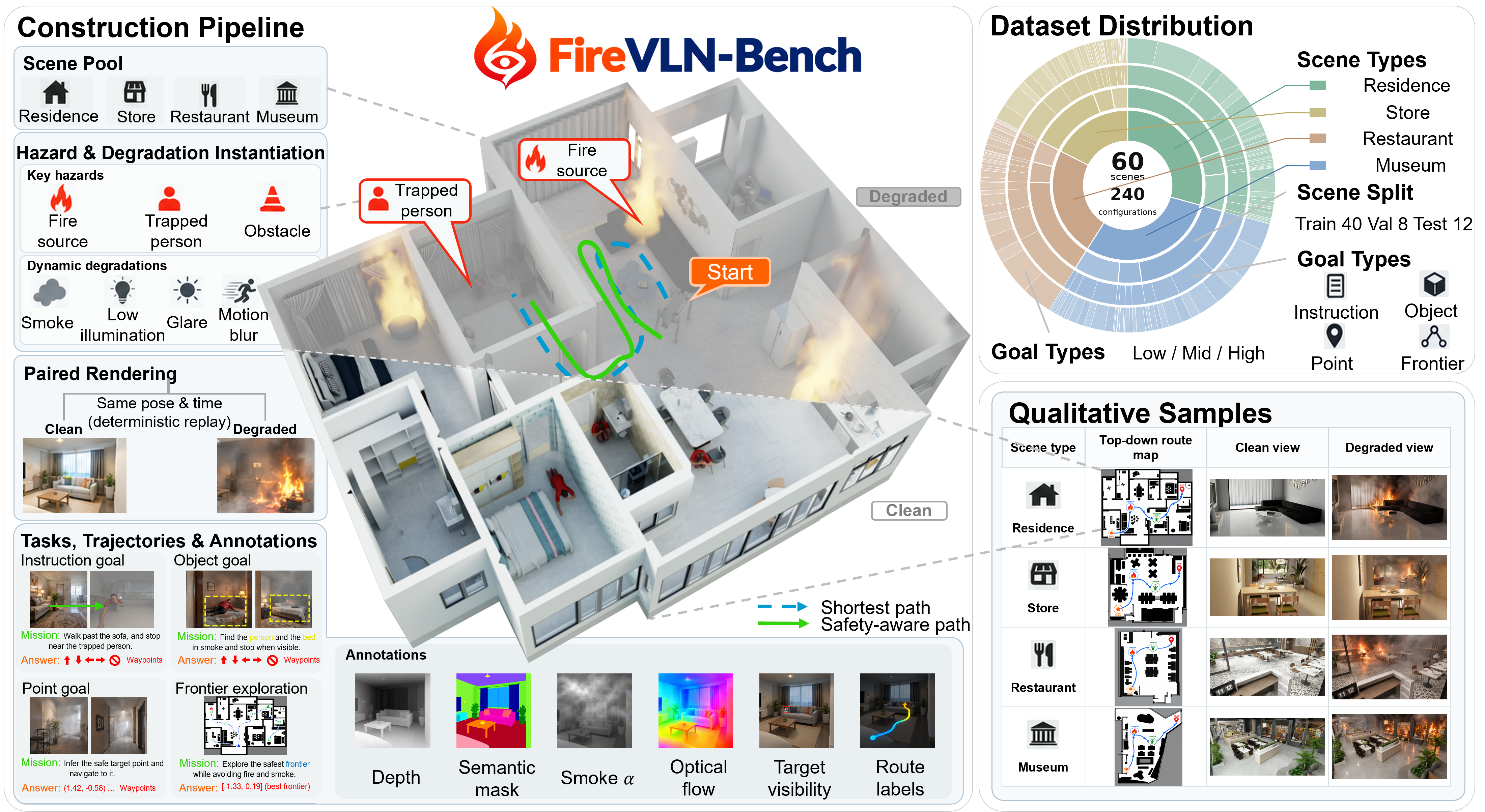}
    \caption{
        Overview of FireVLN-Bench. The figure shows construction pipeline indoor fire scene dataset statistics and qualitative samples.
    }
    \label{fig:firevln-bench}
    \vspace{-5mm}
\end{figure*}

\section{Proposed Benchmark: FireVLN-Bench}
\label{sec:firevln-bench}

\paragraph{Overview.}
To study robust VLN in indoor fire environments, we construct FireVLN-Bench, a paired benchmark that evaluates the effects of visual degradation and reconstruction on perception and navigation in a closed loop. In Isaac Sim, matched settings cover diverse scenes, hazard layouts, and temporal degradations for continuous navigation and robot execution with simulated dynamics. These settings support controlled evaluation of robustness, navigation, safety, and performance recovery through reconstruction. Figure~\ref{fig:firevln-bench} summarizes the benchmark construction pipeline, dataset distribution, and qualitative examples.

\paragraph{Indoor Fire Environment Modeling.}
We denote an environment configuration by $\mathcal E$, comprising scene
geometry, materials, collision assets, and a fixed layout of fire sources,
trapped persons, and task obstacles. For each camera view, degraded and clean
observations at time $t$ are rendered as
\begin{equation}
\begin{aligned}
I_t^{d} &= \mathcal R(\mathcal E,p_t,\mathcal D_t),\\
I_t^{c} &= \mathcal R(\mathcal E,p_t,\mathcal D_t^{c}),
\end{aligned}
\label{eq:paired-fire-rendering}
\end{equation}
where $\mathcal R$ denotes rendering and image formation, $p_t$ is the camera
pose, and $\mathcal D_t$ specifies smoke, illumination, glare, sensor noise,
and motion blur. The reference state $\mathcal D_t^{c}$ suppresses
degradations while retaining fire cues and task semantics. Both observations
share $\mathcal E$, $p_t$, and the fire and object states. This pairing
applies to recorded frames and controlled trajectory replay.

\paragraph{Benchmark Construction.}
FireVLN-Bench is built from indoor scenes curated from SpatialVerse
InteriorAgent and SAGE-3D InteriorGS
USDZ~\citep{interioragent2025,miao2026sage3d}, including residences,
convenience stores, restaurants, and museums. We retain only scenes with valid
collision geometry and tasks that are feasible under clean observations.
InteriorAgent provides editable mesh scenes, while SAGE-3D combines 3D
Gaussian appearance with aligned collision assets.

To simulate fire conditions, we combine each scene with multiple hazard
layouts to form configurations $\mathcal E$. Seeded schedules control fire
intensity and the evolution of $\mathcal D_t$. Using the renderer
$\mathcal R$ in Eq.~\eqref{eq:paired-fire-rendering}, we generate paired
observations $(I_t^d,I_t^c)$ at each recorded pose $p_t$. The clean rendering
state $\mathcal D_t^c$ produces the reference $I_t^c$ for task preservation.
We provide diagnostic images with visual degradation disabled and annotations
including smoke alpha, depth, semantic masks, optical flow, and target
visibility. Each configuration is checked for deterministic replay,
connectivity, collision consistency, and pair alignment. Episodes include a
shortest trajectory and an alternative that accounts for collision risk,
fire proximity, dense smoke, blocked routes, and excessive turning.

\paragraph{Task Formulation and Splits.}
Following continuous and physically embodied VLN
settings~\citep{krantz2020vlnce,wang2025vlnpe}, FireVLN-Bench contains 60
unique indoor scenes with four hazard layouts each, yielding 240
configurations $\mathcal E$. It provides two matched tracks whose canonical
episodes share the environment configuration, task, initial state, and seeded
degradation schedule. FireVLN-Bench-CE evaluates continuous navigation without
teleportation between navigation graph nodes, prior knowledge of the environment
topology, or oracle control of local motion. FireVLN-Bench-PE executes the same
tasks using simulated robot dynamics and a locomotion controller, introducing
contact dynamics, actuation error, camera disturbance, falls, and deadlocks.

The benchmark covers instruction following, object search, point goal
navigation, and frontier exploration. For FireNav, the input image set
$\mathcal I_t$ comprises selected history and current views from the degraded
observation stream $I_t^d$. Each sample contains up to 20 history frames with
their recorded poses $p_t$, current views from the left, front, and right,
and five future local waypoints with heading and arrival labels.
Corresponding references $I_t^c$ support reconstruction evaluation and
comparisons under clean observations. Following protocols for generalization
across environments~\citep{ku2020rxr}, the scenes are split into 40 training
scenes, 8 unseen scenes for validation, and 12 unseen scenes for testing,
corresponding to 160, 32, and 48 hazard configurations. Validation on seen
scenes uses separate episodes from the training environments. Splits are
stratified by scene category and asset source. All configurations
$\mathcal E$ derived from the same scene family remain in the same scene
partition. Clean and degraded pairs, CE and PE counterparts, and nearby
trajectories remain within the same episode split. Episodes involving
extended navigation additionally provide rotational views, candidate
frontiers, memory proxies, subgoals, and labels for safety and information gain.

\begin{figure*}[t]
    \centering
    \includegraphics[width=\textwidth]{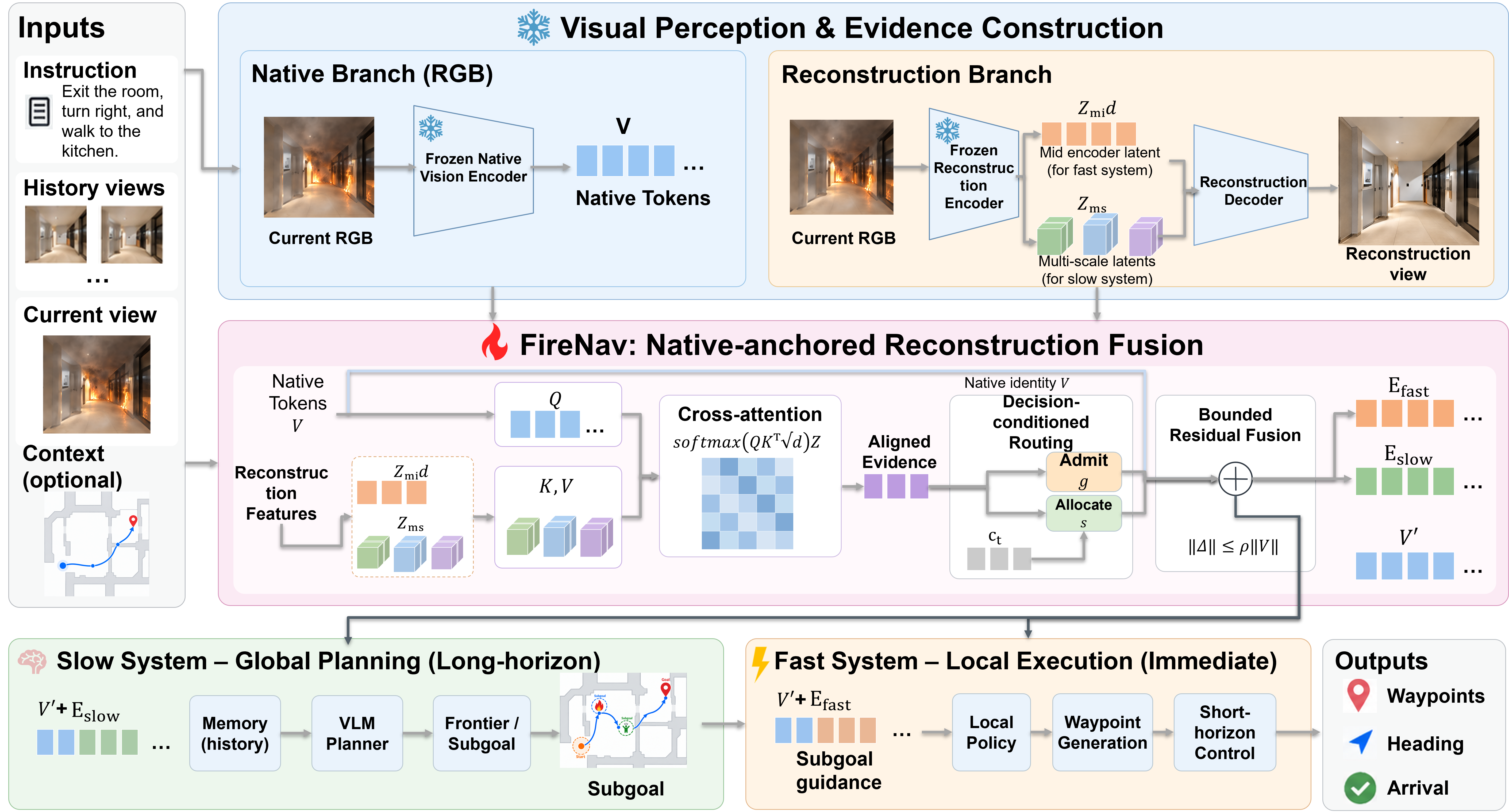}
    \caption{
        Overview of FireNav.
        A frozen native branch provides visual tokens, while a frozen
        reconstruction branch exposes mid-level features for fast execution
        and multiscale features for slow planning.
        FireNav aligns reconstruction features with native tokens, routes them
        according to the current decision state, and injects bounded residual
        evidence while preserving the native visual identity.
        The slow system uses multiscale evidence for memory-based subgoal
        planning, whereas the fast system uses mid-level evidence for local
        waypoint prediction.
    }
    \label{fig:method}
        \vspace{-5mm}
\end{figure*}

\section{Proposed Method: FireNav}
\label{sec:method}

\subsection{Overall Framework}
\label{sec:framework}

As shown in Figure~\ref{fig:method}, FireNav augments a pretrained navigator
with reconstruction evidence while retaining its native visual pathway.
We instantiate the framework on OmniNav~\citep{omninav} and preserve its
visual encoder, vision-language backbone, waypoint interface, and planning
hierarchy. The navigator and reconstruction encoder remain frozen, while the
evidence interface and router are learned, limiting adaptation to auxiliary
representations~\citep{dualvln,maps}. At step $t$, the agent receives a task
condition $\ell_t$ and current and historical images
$\mathcal I_t=\{I_{t,i}\}$ with pose annotations. Intermediate encoder
features support fast execution, and multiscale features support slow planning.
The slow system selects subgoals from memory and candidate frontiers. The
fast system predicts $H=5$ planar waypoints with headings and arrival
probabilities, denoted by $\widehat Y_t$.

\paragraph{Native Token Interface.}
Reconstruction features provide an alternative to using restored RGB as the
navigation input~\citep{unirestore,tasktok}. For image $I_{t,i}$, the native
encoder produces $V_{t,i}\in\mathbb R^{N_i\times d_v}$. An alignment module
projects and pools the reconstruction features into tokens
$Z_{t,i}^{b}\in\mathbb R^{M_i^b\times d_a}$ for branch
$b\in\{\mathrm{fast},\mathrm{slow}\}$. Native tokens retrieve evidence
from the same image through cross attention~\citep{spatialfly,unirestore},
\begin{equation}
Q_{t,i}^{b}=V_{t,i}W_Q^{b},
\qquad
C_{t,i}^{b}=\operatorname{MHA}
\left(Q_{t,i}^{b},Z_{t,i}^{b},Z_{t,i}^{b}\right).
\label{eq:evidence-readout}
\end{equation}
The evidence length $M_i^b$ need not equal $N_i$. The readout occupies the
original visual token slots, leaving text, coordinate tokens, and sequence
length unchanged. Appendix~\ref{app:evidence-interface} details feature
selection and token construction.

\subsection{Decision-Guided Evidence Routing}
\label{sec:evidence-routing}

Adaptive routing allocates visual information according to input demands and
reliability~\citep{ver,x2fusion}. FireNav additionally learns whether that
information improves the navigation prediction. Decision context guides
admission and spatial allocation, utility calibration supervises evidence
selection, and residual control limits the representation change.

\paragraph{Decision Conditioning.}
Navigation depends on task progress and previous actions as well as current
observations~\citep{progressthink,compassnav}. We condition image admission
on visual statistics $q_{t,i}$ and context $c_t$, which includes the task,
an available subgoal, the previous waypoint prediction, and measured execution
progress. Previous waypoints are expressed in the current body frame, and
unavailable context is masked. The context excludes the current prediction,
clean references, and future outcomes. For each branch, the gates are
\begin{equation}
\begin{aligned}
g_{t,i}^{b}
&=\sigma\!\left(f_g^{b}
([\bar Q_{t,i}^{b},\bar C_{t,i}^{b},q_{t,i},c_t])\right),\\
s_{t,i,j}^{b}
&=\sigma\!\left(f_s^{b}
([Q_{t,i,j}^{b},C_{t,i,j}^{b}])\right),
\end{aligned}
\label{eq:evidence-gates}
\end{equation}
where $\sigma$ denotes the sigmoid function, bars indicate token averages,
brackets denote concatenation, and $f_g^b,f_s^b$ are lightweight MLPs.
The scalar $g_{t,i}^b$ sets the contribution of an image, and
$s_{t,i,j}^b$ distributes it across native tokens.

\paragraph{Decision Utility Calibration.}
Decision supervision can use action quality beyond direct
imitation~\citep{compassnav,progressthink}. We supervise admission through
the change in action prediction caused by candidate reconstruction evidence.
A candidate intervention adds spatially weighted reconstruction features to
the native tokens through bounded residual fusion, defined below in
Eqs.~\eqref{eq:candidate-residual}--\eqref{eq:fast-slow-representation}.
The evidence transform and token allocation are first learned with admission
fixed open, then frozen. On a disjoint subset of training episodes, we compare
the native prediction $\widehat Y_t^0$ with $\widehat Y_t^{+i}$, obtained by
admitting the frozen candidate from one view. Other observations, task inputs,
and model states remain unchanged. With the branch index omitted for the
calibrated prediction, the utility is
\begin{equation}
u_{t,i}=\operatorname{sg}\!\left[
\mathcal L_{\mathrm{act}}(\widehat Y_t^0,Y_t^\star)
-\mathcal L_{\mathrm{act}}(\widehat Y_t^{+i},Y_t^\star)
\right],
\label{eq:utility}
\end{equation}
where $\operatorname{sg}$ stops gradients, $Y_t^\star$ denotes the expert
target, and $\mathcal L_{\mathrm{act}}$ combines position, heading, and
arrival errors. Utilities above $\delta>0$ label beneficial evidence,
values below $-\delta$ label harmful evidence, and the neutral interval is
excluded. Binary cross entropy trains the corresponding admission gate.
The candidate comparison is independent of the learned gate, so its current
admission decision cannot determine its own utility label. These comparisons
run only offline and measure isolated view interventions, not additive
contributions or guaranteed episode success. The full action loss and
calibration procedure appear in Appendix~\ref{app:utility-details}.

\paragraph{Bounded Residual Evidence.}
Utility calibration supervises evidence admission, but gate values alone do
not bound the resulting representation change. Small gates can still produce
large updates when projected features have large magnitudes. To limit changes
to native representations during auxiliary fusion~\citep{spatialfly,maps},
we constrain the actual residual,
\begin{align}
\Delta_{t,i}^{b}
&=\gamma_b g_{t,i}^{b}
\big(s_{t,i}^{b}\odot(C_{t,i}^{b}W_O^{b})\big),
\label{eq:candidate-residual}\\
E_{t,i}^{b}
&=\Delta_{t,i}^{b}
\min\!\left(1,
\frac{\rho_b\|V_{t,i}\|_F}{\|\Delta_{t,i}^{b}\|_F+\epsilon}\right),
\label{eq:bounded-intervention}\\
\widetilde V_{t,i}^{b}
&=V'_{t,i}+b_{t,i}E_{t,i}^{b},
\qquad V'_{t,i}\equiv V_{t,i}.
\label{eq:fast-slow-representation}
\end{align}
Here, $\gamma_b$ is a learned scale, $W_O^b$ is the output projection,
$\rho_b$ bounds relative perturbation, $\epsilon>0$ stabilizes normalization,
and $b_{t,i}\in\{0,1\}$ indicates valid evidence.
Token weights are broadcast over channels. Initializing $\gamma_b$ to zero
recovers the native representation at initialization. Invalid evidence
bypasses fusion. The same bounded fusion is used to construct candidate
predictions during calibration and to inject routed evidence during inference.
The constraint bounds feature changes, not action safety. The intervention
ratio and validity checks are given in Appendix~\ref{app:residual-details}.

\subsection{Optimization and Execution}
\label{sec:optimization-execution}

\paragraph{Preservation and Task Learning.}
We train on clean and degraded episodes with task learning and representation
preservation~\citep{maps,dualvln},
\begin{equation}
\mathcal L=
\mathcal L_{\mathrm{nav}}^{(h)}
+\lambda_{\mathrm{ret}}\mathcal L_{\mathrm{ret}}
+\lambda_{\mathrm{int}}\mathcal L_{\mathrm{int}}
+\lambda_u\mathcal L_u.
\label{eq:training-objective}
\end{equation}

The direct head uses the action loss, and the flow head retains its
flow matching objective. Retention penalizes changes on clean states where
native predictions meet expert tolerances. The intervention loss penalizes
residual magnitude. Candidate learning omits $\mathcal L_u$.
Calibration fixes the candidate and optimizes admission with the full
objective. Appendix~\ref{app:optimization-details} details losses
and parameter updates.

\paragraph{Cached Execution.}
Following navigation systems that reuse visual history~\citep{omninav,dualvln},
we cache frozen reconstruction features for their original frames and views.
Routing is recomputed using the current context. The slow system selects a
subgoal with multiscale evidence, and the fast system uses intermediate
features for waypoint prediction. Missing, incompatible, or unfinished
evidence follows the native path. Caching avoids repeated encoding, not the
cost of processing new images. The inherited planning logic and locomotion
controller remain unchanged. Cache keys and scheduling details are provided
in Appendix~\ref{app:cache-details}.

\begin{table*}[t]
    \centering
    \caption{
        Quantitative comparison on FireVLN-Bench unseen test split.
        Among methods with degraded observations, best and second-best
        results are highlighted in bold and underline, respectively.
    }
    \label{tab:main_results}

    \small
    \setlength{\tabcolsep}{3.8pt}
    \renewcommand{\arraystretch}{1.12}

    \begin{tabular}{@{}llcccccccc@{}}
        \toprule
        \multirow[c]{2}{*}{\textbf{Method}}
        & \multirow[c]{2}{*}{\textbf{Venue}}
        & \multicolumn{4}{c}{\textbf{Physical Environment}}
        & \multicolumn{4}{c}{\textbf{Continuous Environment}} \\
        \cmidrule(lr){3-6}
        \cmidrule(l){7-10}
        &
        & NE$\downarrow$ & OS$\uparrow$ & SR$\uparrow$ & SPL$\uparrow$
        & NE$\downarrow$ & OS$\uparrow$ & SR$\uparrow$ & SPL$\uparrow$ \\
        \midrule

        Native VLN (clean)
        & -
        & 4.58 & 68.2 & 54.1 & 42.3
        & 3.32 & 79.7 & 67.5 & 56.6 \\

        Native VLN (degraded)
        & -
        & 7.34 & 43.0 & 26.8 & 17.3
        & 5.94 & 56.3 & 41.8 & 29.9 \\

        \midrule
        \multicolumn{10}{@{}l}{\textit{Reconstructed RGB Frontends}} \\

        AdaIR~\citep{adair}
        & ICLR’25
        & 7.10 & 45.1 & 29.4 & 19.7
        & 5.82 & 57.2 & 42.7 & 30.5 \\

        DCPT~\citep{dcpt}
        & ICLR’25
        & 6.68 & 48.8 & 34.1 & 25.1
        & 5.33 & 62.0 & 47.5 & 36.6 \\

        MoCE-IR-S~\citep{moceir}
        & CVPR’25
        & \underline{6.08} & 50.9 & 34.8 & \underline{27.8}
        & 5.21 & 62.9 & \underline{51.2} & 38.3 \\

        DFPIR~\citep{dfpir}
        & CVPR’25
        & 6.65 & 49.5 & 32.9 & 23.7
        & 5.46 & 60.9 & 46.2 & 34.9 \\

        R2R-Compos~\citep{r2rrestore}
        & CVPR’26
        & 6.52 & 50.5 & 34.5 & 24.8
        & 5.32 & 61.9 & 48.2 & 36.6 \\

        DATPRL-IR~\citep{datprlir}
        & ICLR’26
        & 6.68 & 49.0 & 33.5 & 23.7
        & 5.44 & 60.8 & 47.2 & 35.6 \\

        HOGformer~\citep{hogformer}
        & AAAI’26
        & 6.84 & 47.7 & 31.5 & 22.3
        & 5.57 & 60.0 & 44.5 & 33.1 \\

        BiLaLoRA~\citep{bilalora}
        & CVPR’26
        & 6.34 & \underline{54.2} & 35.6 & 25.2
        & \underline{4.86} & 63.1 & 48.8 & \underline{40.1} \\

        \midrule
        \multicolumn{10}{@{}l}{\textit{Task Adaptation and Reconstruction Evidence}} \\

        Raw RGB + Adapter
        & -
        & 6.35 & 52.4 & 35.3 & 25.2
        & 5.04 & 64.6 & 49.8 & 38.6 \\

        Restored RGB + Adapter
        & -
        & 6.16 & 53.9 & \underline{37.3} & 27.1
        & 4.99 & \underline{66.2} & 50.8 & 39.5 \\

        \textbf{FireNav}
        & \textbf{Ours}
        & \textbf{5.65} & \textbf{59.0}
        & \textbf{41.6} & \textbf{32.0}
        & \textbf{4.53} & \textbf{68.8}
        & \textbf{56.1} & \textbf{45.5} \\

        \bottomrule
    \end{tabular}

    \vspace{-2mm}
\end{table*}

\section{Experiments}
\label{sec:experiments}

\subsection{Experimental Settings}
\label{sec:experimental-settings}

\vspace{-1mm}

\paragraph{Benchmarks and Tasks.}
Experiments use FireVLN-Bench's matched CE and PE tracks in Isaac Sim
(Section~\ref{sec:firevln-bench}). Instruction following on the unseen test
split is the main task. Additional scene splits and goal tasks appear in
Appendix~\ref{app:generalization}. Methods share task specifications,
initial states, hazard layouts, degradation schedules, and episode manifests,
but execute independent trajectories.

\vspace{-1mm}

\paragraph{Metrics.}
We report Navigation Error (NE), Oracle Success (OS),
Success Rate (SR), and Success weighted by Path Length
(SPL)~\citep{omninav,dualvln,wang2025vlnpe}. NE is in meters, and OS, SR,
and SPL are percentages. Action diagnostics report average and final
displacement errors (ADE/FDE), heading error, and arrival F1.
Clean observations are evaluated to assess performance retention.
Physical failures and stopping errors appear in Appendix~\ref{app:safety}.

\vspace{-1mm}

\paragraph{Baselines.}
The RGB frontends are AdaIR~\citep{adair}, DCPT~\citep{dcpt},
MoCE-IR-S~\citep{moceir}, DFPIR~\citep{dfpir},
R2R-Compos~\citep{r2rrestore}, DATPRL-IR~\citep{datprlir},
HOGformer~\citep{hogformer}, and BiLaLoRA~\citep{bilalora}.
Each supplies reconstructed images to the same navigator. Raw RGB + Adapter
and Restored RGB + Adapter add navigation supervision to the corresponding
inputs. The latter uses the same reconstruction model as FireNav.
FireNav uses the R2R-Compos encoder and learns only the evidence interface
and router. Clean and degraded native inputs provide reference conditions.

\vspace{-1mm}

\paragraph{Implementation.}
Comparisons share the navigation checkpoint, visual history, waypoint
interface, controller, and STOP rule. Reconstruction weights remain frozen.
The reader uses width 384 and six attention heads. The main
comparisons use the direct waypoint head. Runtime covers auxiliary encoding
and fusion and reports P50/P95 latency, update frequency, and peak memory.

Closed-loop methods share initial conditions and task definitions but execute
independent trajectories. Paired clean/degraded observations are used only for
same-state comparisons and are excluded from online policy inputs. This avoids
reference leakage during closed-loop navigation.

\subsection{Main Results}
\label{sec:main-results}

Table~\ref{tab:main_results} lists 41.6 SR and 32.0 SPL for FireNav in PE,
and 56.1 SR and 45.5 SPL in CE. Relative to the degraded native baseline,
SR/SPL are higher by 14.8/14.7 points in PE and 14.3/15.6 points in CE.
FireNav has the best values in all eight columns among methods using degraded
observations. Its SPL exceeds the strongest competing value by 4.2 points
in PE and 5.4 points in CE. The clean reference is excluded from this ranking.

Relative to Restored RGB + Adapter, SR/SPL differences are 4.3/4.9 points
in PE and 5.3/6.0 points in CE. NE decreases by 0.51 and 0.46 m.
Both configurations share the reconstruction model but integrate its
output differently, comparing restored RGB with routed
reconstruction features. This isolates the benefit of feature-level evidence
routing beyond reconstruction and task adaptation.

Across methods, PE remains harder than CE, where physical execution introduces
contact dynamics, actuation error, and deadlocks. FireNav maintains its advantage
under challenging fire conditions in both tracks, showing consistent gains beyond
continuous navigation. This suggests that routed reconstruction evidence remains
effective under additional physical execution disturbances.

\begin{figure*}[t]
    \centering
    \includegraphics[width=\textwidth]{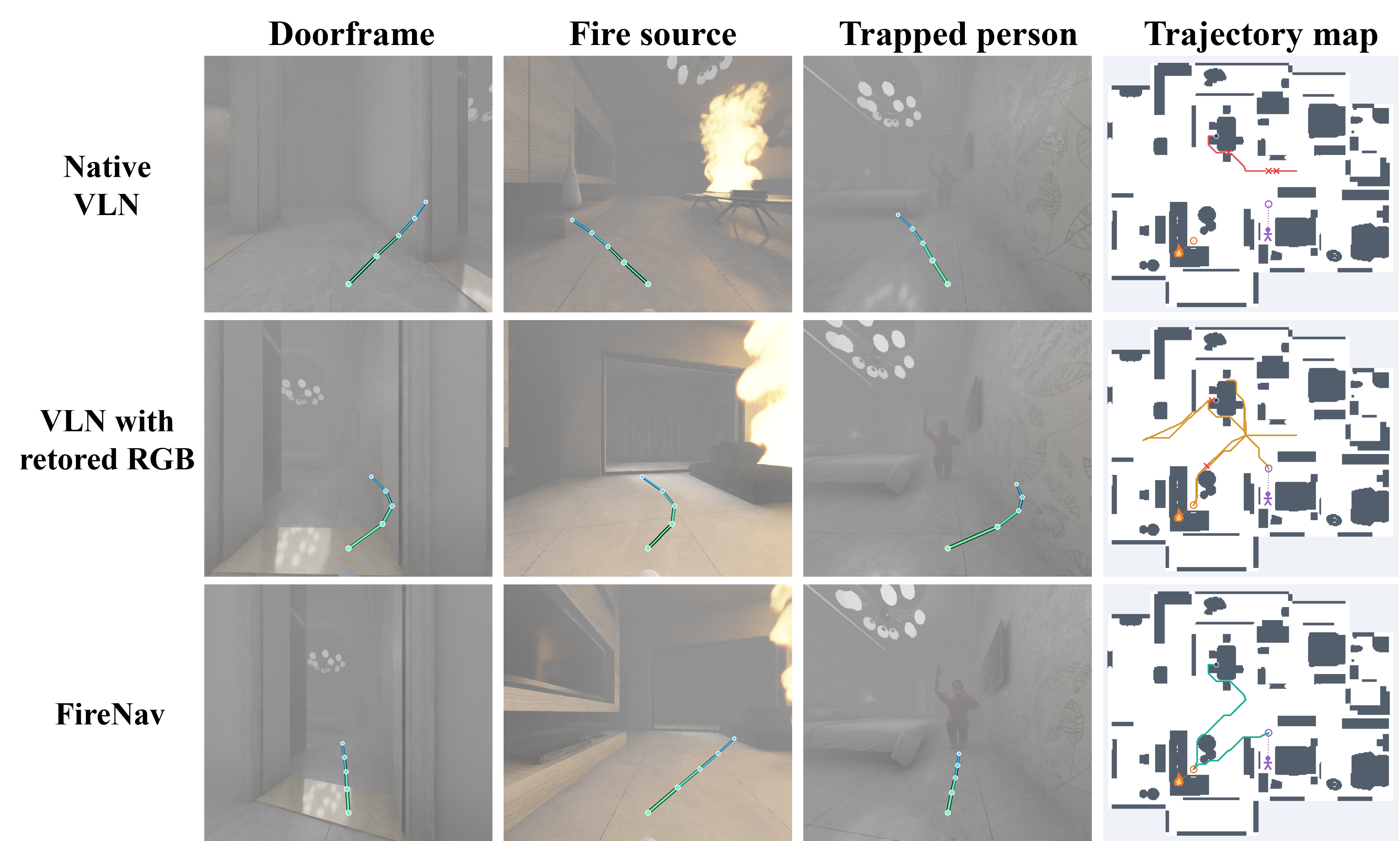}
    \caption{
        Qualitative comparison of waypoint predictions and trajectories.
        Each row shows a model configuration.
        The first three columns show local waypoint predictions at keyframes,
        and the last shows trajectories on occupancy maps.
        Red crosses mark task interruptions (\texttt{STOP}).
    }
    \label{fig:comparison}
\vspace{-3mm}
\end{figure*}

Figure~\ref{fig:comparison} complements the aggregate results with local
predictions and trajectory views. The selected keyframes cover a doorway,
a fire source, and a person under smoke occlusion. Waypoint overlays expose
differences in local direction, while the maps show route choices and
interruption locations. The keyframes depict different camera positions
along the respective trajectories rather than a controlled comparison
at an identical navigation state. The visualization further highlights how
local prediction differences accumulate into distinct navigation outcomes.

\subsection{Analysis of Decision-Guided Evidence Routing}
\label{sec:routing-analysis}

\begin{table*}[t]
    \centering
    \caption{
        Ablation study of FireNav.
        Clean SR measures retention of the native behavior under clean
        observations. All variants use the same navigation backbone and
        evaluation protocol.
    }
    \label{tab:ablation_components}

    \vspace{-1mm}

    \small
    \setlength{\tabcolsep}{6.0pt}
    \renewcommand{\arraystretch}{1.12}

    \begin{tabular}{@{}lcccccc@{}}
        \toprule
        \multirow[c]{2}{*}{\textbf{Variant}}
        & \multicolumn{3}{c}{\textbf{Physical Environment}}
        & \multicolumn{3}{c}{\textbf{Continuous Environment}} \\
        \cmidrule(lr){2-4}
        \cmidrule(l){5-7}
        & SR$\uparrow$ & SPL$\uparrow$ & Clean SR$\uparrow$
        & SR$\uparrow$ & SPL$\uparrow$ & Clean SR$\uparrow$ \\
        \midrule

        \textbf{Full FireNav}
        & \textbf{41.6} & \textbf{32.0} & 52.8
        & \textbf{56.1} & \textbf{45.5} & 67.1 \\

        Always-on evidence
        & 36.1 & 26.0 & 50.4
        & 50.5 & 38.6 & 65.6 \\

        w/o decision context
        & 38.9 & 29.0 & 53.2
        & 53.7 & 42.1 & 67.1 \\

        w/o quality statistics
        & 39.8 & 30.0 & 53.6
        & 54.4 & 43.1 & 66.8 \\

        w/o utility calibration
        & 37.8 & 28.0 & 52.9
        & 52.2 & 40.6 & 66.7 \\

        w/o residual control
        & 39.9 & 29.7 & 51.6
        & 53.2 & 41.9 & 66.2 \\

        w/o clean retention
        & 42.8 & 33.0 & 48.6
        & 57.2 & 46.3 & 63.7 \\

        Image gate only
        & 40.8 & 30.8 & 53.8
        & 55.4 & 44.4 & 67.8 \\

        Random encoder + trained interface
        & 37.8 & 27.5 & 52.8
        & 51.6 & 39.7 & 67.0 \\

        \bottomrule
    \end{tabular}
    
\vspace{-1mm}
\end{table*}

\paragraph{Component Analysis.}
In Table~\ref{tab:ablation_components}, unconditional fusion has 6.0/6.9
points lower PE/CE SPL than the full configuration. Removing decision
context lowers SPL by 3.0/3.4 points, and removing utility calibration
lowers it by 4.0/4.9 points. Without residual control, SPL decreases by
2.3/3.6 points. Image gating alone loses 1.2/1.1 points.
Removing clean retention raises degraded SPL by 1.0/0.8 points but lowers
clean SR by 4.2/3.4 points. The full configuration balances degraded
performance and clean retention rather than maximizing every column.

\vspace{-3mm}

\paragraph{Evidence Content.}
Appendix~\ref{app:interventions} distinguishes trained variants from input
interventions on one checkpoint. Shuffled content has 13.0/15.1 points lower
PE/CE SPL than matched evidence. Shuffling only the router context reduces
SPL by 3.3/4.5 points. These comparisons examine dependence on evidence
content and decision context. The random encoder and adapter configurations
provide separate controls for reconstruction pretraining and task adaptation.

\vspace{-3mm}

\subsection{Action Quality and Efficiency}
\label{sec:action-efficiency}

\paragraph{Action Prediction.}
Table~\ref{tab:action_diagnostics} in Appendix~\ref{app:action-diagnostics}
lists degraded ADE/FDE of 0.504/0.861 m for FireNav, compared with
0.555/0.926 m for Restored RGB + Adapter. Heading error is
$15.2^\circ$ versus $17.0^\circ$, and arrival F1 is 0.667 versus 0.586.
Clean ADE remains close to the native value, at 0.506 versus 0.504 m.
These local diagnostics are distinct from episode success and STOP outcomes
.
\vspace{-3mm}

\begin{table*}[t]
    \centering
    \caption{
        Runtime comparison under the same execution setting. P50/P95 denote update latency.
    }
    \label{tab:runtime}

    \vspace{-1mm}

    \small
    \setlength{\tabcolsep}{5.0pt}
    \renewcommand{\arraystretch}{1.12}

    \resizebox{\textwidth}{!}{%
    \begin{tabular}{@{}lcccccc@{}}
        \toprule
        \multirow[c]{2}{*}{\textbf{Method}}
        & \multicolumn{2}{c}{\textbf{Additional Cost}}
        & \multicolumn{2}{c}{\textbf{End-to-End Latency}}
        & \multicolumn{2}{c}{\textbf{System}} \\
        \cmidrule(lr){2-3}
        \cmidrule(lr){4-5}
        \cmidrule(l){6-7}
        & Encoder/Restoration (ms)$\downarrow$
        & Fusion (ms)$\downarrow$
        & P50 (ms)$\downarrow$
        & P95 (ms)$\downarrow$
        & Hz$\uparrow$
        & Peak GB$\downarrow$ \\
        \midrule

        Native VLN
        & 0.0 & 0.0 & \textbf{280.1} & \textbf{359.2}
        & \textbf{3.53} & \textbf{18.06} \\

        AdaIR
        & 209.9 & 2.0 & 491.9 & 597.4 & 2.02 & 20.26 \\

        DCPT
        & 96.5 & 2.0 & 378.8 & 464.8 & 2.62 & 19.16 \\

        MoCE-IR-S
        & 126.1 & 2.0 & 408.5 & 498.1 & 2.43 & 19.36 \\

        DFPIR
        & 113.9 & 2.0 & 396.9 & 485.0 & 2.51 & 19.66 \\

        R2R-Compos
        & 78.3 & 2.0 & 361.2 & 445.0 & 2.75 & 19.06 \\

        DATPRL-IR
        & 134.8 & 2.0 & 417.3 & 508.5 & 2.38 & 19.86 \\

        HOGformer
        & 270.4 & 2.0 & 551.3 & 668.4 & 1.80 & 20.56 \\

        BiLaLoRA
        & 71.8 & 2.0 & 354.0 & 436.5 & 2.80 & 19.56 \\

        Raw RGB + Adapter
        & 0.0 & 8.0 & 288.0 & 367.3 & 3.43 & 18.46 \\

        Restored RGB + Adapter
        & 77.9 & 10.0 & 368.7 & 452.4 & 2.69 & 19.46 \\

        \textbf{FireNav}
        & 17.9 & 12.0 & 310.1 & 389.4 & 3.19 & 19.16 \\

        \bottomrule
    \end{tabular}%
    }

    \vspace{-4mm}
\end{table*}

\paragraph{Runtime.}
Table~\ref{tab:runtime} lists 17.9 ms for reconstruction feature encoding
and 12.0 ms for fusion. FireNav's P95 latency is 389.4 ms,
8.4\% above the native baseline and 63.0 ms below Restored RGB + Adapter.
The update frequencies are 3.19, 3.53, and 2.69 Hz.
Direct use of encoder features avoids RGB decoding on the fast path.
Appendix~\ref{app:history} separates history length from feature reuse.
These latency figures describe navigation updates rather than the frequency
of the locomotion controller.

\vspace{-3mm}

\section{Conclusion}
\label{sec:conclusion}

\vspace{-3mm}

We have presented FireNav, a decision-oriented reconstruction evidence routing framework for robust VLN in indoor fire environments. 
To advance benchmarking research in this field, FireVLN-Bench is first introduced as a paired benchmark with clean references, safety supervision, and matched settings for continuous navigation and robot execution.
Our method preserves native visual tokens while selectively incorporating reconstruction evidence through lightweight routing and decision utility calibration. 
Extensive experiments demonstrate consistent improvements over state-of-the-art baselines in navigation success and path efficiency across evaluation settings.

\vspace{-3mm}

\bibliography{references}
\bibliographystyle{iclr2027_conference}

\appendix

\section{Additional Method Details}
\label{app:method-details}

\subsection{Evidence Sources and Token Construction}
\label{app:evidence-interface}

The reconstruction encoder provides an intermediate feature
$R_{t,i}^{\mathrm{mid}}$ and multiscale features
$\{R_{t,i}^{l}\}_{l\in\mathcal L}$. The evidence source for branch $b$ is
\begin{equation}
\mathcal R_{t,i}^{b}=
\begin{cases}
R_{t,i}^{\mathrm{mid}}, & b=\mathrm{fast},\\
\{R_{t,i}^{l}\}_{l\in\mathcal L}, & b=\mathrm{slow}.
\end{cases}
\label{eq:branch-features}
\end{equation}
Scale projections and spatial pooling construct the evidence tokens,
\begin{equation}
Z_{t,i}^{b}=A_\phi^{b}
\left(\mathcal R_{t,i}^{b},q_{t,i},\nu_i\right)
\in\mathbb R^{M_i^b\times d_a}.
\label{eq:evidence-tokens}
\end{equation}
Here $q_{t,i}$ contains brightness, contrast, saturation, and gradient
statistics, and $\nu_i$ identifies the view. Cross attention returns $N_i$
tokens. The output projection restores the native width $d_v$ before
residual addition. Encoder evidence does not require RGB decoding on the
fast path.

\subsection{Residual Validity and Initialization}
\label{app:residual-details}

The validity flag $b_{t,i}$ differs from the branch index $b$. Frame identity,
preprocessing, checkpoint consistency, and finite tensor values are checked
before fusion. Invalid evidence returns the native tokens directly, rather
than multiplying invalid values by zero. The notation $V'_{t,i}\equiv V_{t,i}$
in Figure~\ref{fig:method} denotes the identity stream. Evidence is added once.
Initializing $\gamma_b=0$ gives the native representation. For
$\rho_b\geq0$ and $\epsilon>0$,
\begin{equation}
r_{t,i}^{b}=
\frac{\|b_{t,i}E_{t,i}^{b}\|_F}{\|V_{t,i}\|_F+\epsilon}
\leq\rho_b.
\label{eq:intervention-ratio}
\end{equation}
The bound follows from Eq.~\eqref{eq:bounded-intervention} and applies to the
whole image representation. It is not a bound on each token or on action error.

\subsection{Action Loss and Utility Calibration}
\label{app:utility-details}

For predicted positions $\hat p_j$, headings $\hat\theta_j$, and arrival
probabilities $\hat a_j$, with expert targets marked by $\star$, the action
loss is
\begin{equation}
\begin{aligned}
\mathcal L_{\mathrm{act}}(\widehat Y,Y^\star)
=\frac{1}{H}\sum_{j=1}^{H}\Big[&
\lambda_p\operatorname{SmoothL1}(\hat p_j-p_j^\star)\\
&+\lambda_\theta\big(1-\cos(\hat\theta_j-\theta_j^\star)\big)\\
&+\lambda_a\operatorname{BCE}(\hat a_j,a_j^\star)\Big].
\end{aligned}
\label{eq:action-loss}
\end{equation}
Coordinates follow the native decoder and normalization convention.
Calibration compares the native prediction with a frozen candidate admitted
from one view. All other observations, task inputs, and model states are fixed.
The candidate admission is open when constructing the utility label. It does
not depend on the learned gate. Values within $[-\delta,\delta]$ are excluded,
\begin{equation}
\mathcal L_u=
\mathbb E_{|u_{t,i}|>\delta}
\operatorname{BCE}\!\left(g_{t,i},\mathbf 1[u_{t,i}>\delta]\right).
\label{eq:utility-calibration}
\end{equation}
The branch index is suppressed for the calibrated prediction. Calibration
episodes belong to the training split and are excluded from candidate learning.
The labels describe isolated view interventions. Predictions using multiple
views are also supervised by the navigation objective.

\subsection{Loss Terms and Training Stages}
\label{app:optimization-details}

The direct head uses Eq.~\eqref{eq:action-loss}. The Method also describes
retaining the native objective for a flow head~\citep{omninav,dualvln},
which requires its own checkpoint and decoding convention. No flow-head
results are included in this revision. Clean retention compares assisted
outputs with detached native predictions only where the latter meet expert
tolerances. It uses position, circular heading, and arrival probability
differences. The intervention term is
\begin{equation}
\mathcal L_{\mathrm{int}}=
\mathbb E_t\left[
\frac{1}{|\mathcal B_t|\,|\mathcal I_t|}
\sum_{b\in\mathcal B_t}\sum_i(r_{t,i}^{b})^2
\right],
\label{eq:intervention-loss}
\end{equation}
where $\mathcal B_t$ contains the active branches. Candidate learning fixes
admission open and trains the evidence transform and token allocation without
$\mathcal L_u$. Calibration freezes the candidate and trains admission using
Eq.~\eqref{eq:training-objective}. Gradients pass through the frozen navigator
without updating its weights.

\subsection{Cache Consistency and Scheduling}
\label{app:cache-details}

Cache keys contain frame identity, view, preprocessing transform, checkpoint,
and precision. Only frozen reconstruction features are reused. Gates are
recomputed from the current context, and historical features remain attached
to their original observations. Missing, incompatible, or unfinished evidence
uses the native path. A pending feature may later serve its matching
historical frame. Reuse does not avoid encoding new observations or change
the inherited planner and locomotion controller.

\section{Additional Experimental Details}
\label{app:experimental-details}

\vspace{-3mm}

\subsection{Benchmark Allocation}
\label{app:benchmark-allocation}

\vspace{-3mm}

\begin{table*}[h]
    \centering
    \caption{
    Data allocation in FireVLN-Bench. Validation seen uses held-out
    episodes from the training scenes and therefore does not introduce
    additional scene identities.}
    \label{tab:benchmark_allocation}

    \small
    \setlength{\tabcolsep}{4pt}
    \renewcommand{\arraystretch}{1.12}

    \resizebox{\textwidth}{!}{%
    \begin{tabular}{@{}lccccc@{}}
        \toprule
        \textbf{Split} & \textbf{Scenes} & \textbf{Layouts}
        & \textbf{Routes/Layout/Task} & \textbf{Episodes/Task}
        & \textbf{Four-Task Episodes} \\
        \midrule
        Train              & 40 & 160 & 100 & 16,000 & 64,000 \\
        Validation Seen    & 40 & 160 & 25  & 4,000  & 16,000 \\
        Validation Unseen  & 8  & 32  & 25  & 800    & 3,200 \\
        Test Unseen        & 12 & 48  & 25  & 1,200  & 4,800 \\
        \bottomrule
    \end{tabular}%
    }
    \vspace{-2mm}
\end{table*}

Table~\ref{tab:benchmark_allocation} specifies 40 training, 8 unseen
validation, and 12 unseen test scenes. Seen validation reuses training scenes
but reserves separate episodes. The listed instruction following allocation
is 1,200 test episodes per track. CE and PE are separate executions of
canonical tasks, not additional scene identities.

\subsection{Evaluation Protocol}
\label{app:evaluation-protocol}

Closed loop comparisons share initial conditions and task definitions, not
all subsequent observations. Each policy generates its own trajectory.
Paired clean and degraded frames instead share a camera pose and scene state
for controlled image or prediction comparisons. Action diagnostics and
arrival classification are separate from episode metrics and accepted STOP
events. Clean SR measures task performance, not equality of individual
predictions. Collision, fall, and stuck events need not be mutually exclusive.

\section{Extended Routing and Action Analysis}
\label{app:routing-analysis}

\subsection{Encoder and Routing Strategy}
\label{app:encoder-routing}

\begin{table*}[h]
    \centering
    \caption{
        Factorial comparison of reconstruction encoders and evidence routing
        strategies. Values are SPL (\%). Each encoder uses the same evidence
        budget and optimization recipe.
    }
    \label{tab:encoder_routing}

    \vspace{3.5mm}

    \small
    \setlength{\tabcolsep}{6.8pt}
    \renewcommand{\arraystretch}{1.12}

    \begin{tabular}{@{}lcccccc@{}}
        \toprule
        \multirow[c]{2}{*}{\textbf{Encoder}}
        & \multicolumn{3}{c}{\textbf{Physical Environment}}
        & \multicolumn{3}{c}{\textbf{Continuous Environment}} \\
        \cmidrule(lr){2-4}
        \cmidrule(l){5-7}
        & Always-on
        & Quality Gate
        & FireNav
        & Always-on
        & Quality Gate
        & FireNav \\
        \midrule

        AdaIR
        & 23.6 & 26.0 & \textbf{27.8}
        & 35.6 & 38.0 & \textbf{40.3} \\

        DCPT
        & 27.0 & 28.6 & \textbf{32.5}
        & 39.2 & 41.2 & \textbf{46.0} \\

        MoCE-IR-S
        & 26.4 & 29.4 & \textbf{32.3}
        & 39.1 & 42.1 & \textbf{45.7} \\

        DFPIR
        & 25.4 & 28.6 & \textbf{30.8}
        & 37.9 & 41.2 & \textbf{44.2} \\

        R2R-Compos
        & 26.0 & 29.2 & \textbf{32.0}
        & 38.6 & 41.9 & \textbf{45.5} \\

        DATPRL-IR
        & 26.0 & 28.9 & \textbf{31.7}
        & 38.3 & 41.9 & \textbf{45.4} \\

        HOGformer
        & 25.4 & 27.8 & \textbf{30.2}
        & 37.0 & 40.3 & \textbf{43.9} \\

        BiLaLoRA
        & 26.7 & 28.5 & \textbf{31.3}
        & 38.2 & 41.4 & \textbf{44.9} \\

        \bottomrule
    \end{tabular}

    \vspace{-2mm}
\end{table*}

Within each encoder in Table~\ref{tab:encoder_routing}, FireNav routing has
higher SPL than unconditional fusion or quality gating. R2R-Compos gives
32.0/45.5 PE/CE SPL, compared with 32.5/46.0 for DCPT. Encoder choice and
routing strategy are distinct factors. This table alone does not explain why
R2R-Compos is selected.

\subsection{Evidence Content Interventions}
\label{app:interventions}

\begin{table*}[h]
    \centering
    \caption{
        Evidence intervention analysis with the same trained FireNav
        checkpoint. Evidence content or router context is modified without
        retraining the navigator.
    }
    \label{tab:evidence_interventions}

    \vspace{3.5mm}

    \small
    \setlength{\tabcolsep}{5.8pt}
    \renewcommand{\arraystretch}{1.12}

    \begin{tabular}{@{}lcccccc@{}}
        \toprule
        \multirow[c]{2}{*}{\textbf{Intervention}}
        & \multicolumn{3}{c}{\textbf{Physical Environment}}
        & \multicolumn{3}{c}{\textbf{Continuous Environment}} \\
        \cmidrule(lr){2-4}
        \cmidrule(l){5-7}
        & SR$\uparrow$ & SPL$\uparrow$ & Clean SR$\uparrow$
        & SR$\uparrow$ & SPL$\uparrow$ & Clean SR$\uparrow$ \\
        \midrule

        \textbf{Matched evidence}
        & \textbf{41.6} & \textbf{32.0} & 52.8
        & \textbf{56.1} & \textbf{45.5} & 67.1 \\

        Exact bypass
        & 26.8 & 17.3 & 54.1
        & 41.8 & 29.9 & 67.5 \\

        Zero spatial features
        & 28.0 & 18.0 & 54.0
        & 41.2 & 29.5 & 68.1 \\

        Shuffled content
        & 29.3 & 19.0 & 53.6
        & 42.4 & 30.4 & 66.9 \\

        Incorrect view content
        & 27.3 & 17.4 & 53.4
        & 40.5 & 29.2 & 67.3 \\

        Incorrect time content
        & 26.9 & 17.1 & 53.3
        & 40.0 & 28.6 & 67.2 \\

        Shuffled router context only
        & 38.2 & 28.7 & 53.8
        & 52.2 & 41.0 & 67.4 \\

        Invalid metadata / bypass
        & 26.8 & 17.3 & 54.1
        & 41.8 & 29.9 & 67.5 \\

        \bottomrule
    \end{tabular}

\end{table*}

The exact and invalid-metadata bypass rows in
Table~\ref{tab:evidence_interventions} match the native reference.
Shuffled content has 19.0/30.4 PE/CE SPL, compared with 32.0/45.5 for
matched evidence. Incorrect views or times also reduce the listed scores.
These are interventions on one checkpoint, not independently trained ablations.
The injection point must distinguish a valid packet containing altered content
from an invalid packet rejected by the bypass checks.

\subsection{Local Action Diagnostics}
\label{app:action-diagnostics}

\begin{table*}[h]
    \centering
    \caption{
        Action prediction under degraded and clean observations.
        ADE/FDE are measured in meters, heading error in degrees, and
        arrival F1 as a fraction.
    }
    \label{tab:action_diagnostics}

    \vspace{3.5mm}

    \small
    \setlength{\tabcolsep}{5.1pt}
    \renewcommand{\arraystretch}{1.12}

    \begin{tabular}{@{}lccccccc@{}}
        \toprule
        \multirow[c]{2}{*}{\textbf{Method}}
        & \multicolumn{4}{c}{\textbf{Degraded Observations}}
        & \multicolumn{3}{c}{\textbf{Clean Observations}} \\
        \cmidrule(lr){2-5}
        \cmidrule(l){6-8}
        & ADE$\downarrow$ & FDE$\downarrow$
        & Heading$\downarrow$ & Arrival F1$\uparrow$
        & ADE$\downarrow$ & Heading$\downarrow$ & Arrival F1$\uparrow$ \\
        \midrule

        Native VLN
        & 0.666 & 1.101 & 21.0 & 0.474
        & \textbf{0.504} & \textbf{17.0} & \textbf{0.670} \\

        MoCE-IR-S
        & 0.595 & 0.991 & 18.8 & 0.546
        & 0.516 & 17.5 & 0.657 \\

        R2R-Compos
        & 0.615 & 1.021 & 19.5 & 0.523
        & 0.522 & 17.8 & 0.644 \\

        HOGformer
        & 0.615 & 1.036 & 19.8 & 0.509
        & 0.511 & 17.3 & 0.663 \\

        BiLaLoRA
        & 0.600 & 1.001 & 18.6 & 0.552
        & 0.514 & 17.4 & 0.663 \\

        Raw RGB + Adapter
        & 0.565 & 0.956 & 17.4 & 0.575
        & 0.507 & 17.2 & 0.667 \\

        Restored RGB + Adapter
        & 0.555 & 0.926 & 17.0 & 0.586
        & 0.509 & 17.4 & 0.663 \\

        \textbf{FireNav}
        & \textbf{0.504} & \textbf{0.861}
        & \textbf{15.2} & \textbf{0.667}
        & 0.506 & 17.1 & 0.669 \\

        \bottomrule
    \end{tabular}

    \vspace{-2mm}
\end{table*}

Table~\ref{tab:action_diagnostics} separates local prediction errors from
closed loop results. Relative to Restored RGB + Adapter, FireNav has
$1.8^\circ$ lower degraded heading error and 0.081 higher arrival F1.
Clean ADE is 0.506 m, compared with 0.504 m for the native model.
An arrival classification score is not a rate of successful episode termination.

\subsection{Image Quality and Action Utility}
\label{app:rgb-action}

\begin{table*}[h]
    \centering
    \caption{
        Relationship between image reconstruction quality, action prediction,
        and RGB reconstruction cost.
    }
    \label{tab:rgb_action}

    \vspace{3.5mm}

    \small
    \setlength{\tabcolsep}{8pt}
    \renewcommand{\arraystretch}{1.12}

    \begin{tabular}{@{}lccccc@{}}
        \toprule
        \multirow[c]{2}{*}{\textbf{Method}}
        & \multicolumn{2}{c}{\textbf{Image Quality}}
        & \multicolumn{2}{c}{\textbf{Action ADE}}
        & \textbf{Cost} \\
        \cmidrule(lr){2-3}
        \cmidrule(lr){4-5}
        \cmidrule(l){6-6}
        & PSNR$\uparrow$
        & SSIM$\uparrow$
        & Degraded$\downarrow$
        & Clean$\downarrow$
        & RGB ms$\downarrow$ \\
        \midrule

        Raw degraded
        & 16.8 & 0.680 & 0.666 & \textbf{0.504} & 0.0 \\

        AdaIR
        & 15.9 & 0.648 & 0.645 & 0.533 & 69.9 \\

        DCPT
        & 17.8 & 0.717 & 0.603 & 0.518 & 31.8 \\

        MoCE-IR-S
        & 18.2 & 0.728 & \textbf{0.595} & 0.516 & 42.0 \\

        DFPIR
        & 17.6 & 0.711 & 0.608 & 0.520 & 38.1 \\

        R2R-Compos
        & 16.2 & 0.658 & 0.615 & 0.522 & 26.1 \\

        DATPRL-IR
        & 17.5 & 0.707 & 0.610 & 0.519 & 44.9 \\

        HOGformer
        & 16.6 & 0.697 & 0.615 & 0.511 & 90.4 \\

        BiLaLoRA
        & \textbf{18.6} & \textbf{0.745} & 0.600 & 0.514 & 24.0 \\

        \bottomrule
    \end{tabular}

    \vspace{-2mm}
\end{table*}

BiLaLoRA has the highest PSNR and SSIM among the RGB frontends in
Table~\ref{tab:rgb_action}, while MoCE-IR-S has the lowest degraded ADE.
R2R-Compos has lower ADE than raw input despite lower PSNR and SSIM.
The listed rankings differ. They do not establish the correlation between
image quality and navigation utility across individual states.

\section{Physical Failures and Stopping}
\label{app:safety}

\begin{table*}[h]
    \centering
    \caption{
        Safety and terminal behavior on FireVLN-Bench.
        All values are percentages of test episodes.
    }
    \label{tab:safety}

    \small
    \setlength{\tabcolsep}{5.4pt}
    \renewcommand{\arraystretch}{1.12}

    \begin{tabular}{@{}lcccccc@{}}
        \toprule
        \multirow[c]{2}{*}{\textbf{Method}}
        & \multicolumn{4}{c}{\textbf{Physical Environment}}
        & \multicolumn{2}{c}{\textbf{Continuous Environment}} \\
        \cmidrule(lr){2-5}
        \cmidrule(l){6-7}
        & Collision$\downarrow$
        & Fall$\downarrow$
        & Stuck$\downarrow$
        & False STOP$\downarrow$
        & Collision$\downarrow$
        & False STOP$\downarrow$ \\
        \midrule

        Native VLN (clean)
        & 10.3 & 1.9 & 6.3 & 3.2 & 3.5 & 3.3 \\

        Native VLN (degraded)
        & 33.2 & 3.2 & 16.2 & 13.5 & 19.8 & 14.5 \\

        AdaIR
        & 31.0 & 3.2 & 15.7 & 12.7 & 18.1 & 13.2 \\

        DCPT
        & 23.6 & 2.2 & 13.7 & 10.0 & 13.2 & 10.2 \\

        MoCE-IR-S
        & 22.5 & 2.1 & 12.8 & 9.0 & 11.8 & 10.1 \\

        DFPIR
        & 24.8 & \textbf{1.4} & 13.8 & 9.8 & 13.8 & 10.8 \\

        R2R-Compos
        & 24.8 & 2.7 & 13.7 & 9.8 & 13.8 & 10.5 \\

        DATPRL-IR
        & 25.3 & 2.8 & 13.9 & 10.6 & 14.1 & 11.3 \\

        HOGformer
        & 27.2 & 2.8 & 14.5 & 11.3 & 15.8 & 11.8 \\

        BiLaLoRA
        & 23.2 & 2.7 & 12.8 & 9.3 & 12.9 & 9.5 \\

        Raw RGB + Adapter
        & 21.8 & 2.1 & 12.3 & 7.7 & 11.0 & 9.2 \\

        Restored RGB + Adapter
        & 20.6 & 2.1 & 11.7 & 7.2 & 9.9 & 7.8 \\

        \textbf{FireNav}
        & \textbf{16.9} & 2.2
        & \textbf{9.3} & \textbf{5.0}
        & \textbf{7.2} & \textbf{5.7} \\

        \bottomrule
    \end{tabular}

\vspace{3mm}
\end{table*}

Among the degraded configurations in Table~\ref{tab:safety}, FireNav has the
lowest listed collision, stuck, and false STOP rates. Relative to Restored
RGB + Adapter, PE collision and stuck rates are lower by 3.7 and 2.4 points.
Its fall rate is 2.2\%, compared with 2.1\% for the adapter and 1.4\% for
DFPIR. These categories do not support an advantage on every failure mode
or a claim about cumulative smoke exposure.

\section{Performance Across Splits and Tasks}
\label{app:generalization}

\subsection{Seen and Unseen Environments}
\label{app:scene-generalization}

\begin{table*}[h]
    \centering
    \caption{
        Generalization across seen validation, unseen validation, and unseen
        test environments.
    }
    \label{tab:split_generalization}

    \small
    \setlength{\tabcolsep}{6.2pt}
    \renewcommand{\arraystretch}{1.12}

    \begin{tabular}{@{}lcccc@{}}
        \toprule
        \multirow[c]{2}{*}{\textbf{Method / Split}}
        & \multicolumn{2}{c}{\textbf{Physical Environment}}
        & \multicolumn{2}{c}{\textbf{Continuous Environment}} \\
        \cmidrule(lr){2-3}
        \cmidrule(l){4-5}
        & SR$\uparrow$ & SPL$\uparrow$
        & SR$\uparrow$ & SPL$\uparrow$ \\
        \midrule

        Native VLN / Seen Val.
        & 36.4 & 26.2 & 45.8 & 35.0 \\
        Native VLN / Unseen Val.
        & 33.5 & 22.6 & 40.0 & 29.5 \\
        Native VLN / Unseen Test
        & 26.8 & 17.3 & 41.8 & 29.9 \\

        \midrule

        BiLaLoRA / Seen Val.
        & 44.7 & 34.0 & 53.8 & 43.0 \\
        BiLaLoRA / Unseen Val.
        & 40.6 & 29.6 & 48.0 & 37.4 \\
        BiLaLoRA / Unseen Test
        & 35.6 & 25.2 & 48.8 & 40.1 \\

        \midrule

        Restored RGB + Adapter / Seen Val.
        & 47.0 & 36.4 & 56.5 & 45.6 \\
        Restored RGB + Adapter / Unseen Val.
        & 42.4 & 31.4 & 50.5 & 39.8 \\
        Restored RGB + Adapter / Unseen Test
        & 37.3 & 27.1 & 50.8 & 39.5 \\

        \midrule

        FireNav / Seen Val.
        & \textbf{51.8} & \textbf{41.7}
        & \textbf{63.0} & \textbf{52.5} \\
        FireNav / Unseen Val.
        & \textbf{46.9} & \textbf{36.6}
        & \textbf{56.6} & \textbf{46.3} \\
        FireNav / Unseen Test
        & \textbf{41.6} & \textbf{32.0}
        & \textbf{56.1} & \textbf{45.5} \\

        \bottomrule
    \end{tabular}

    \vspace{-2mm}
\end{table*}

FireNav exceeds Restored RGB + Adapter on each split in
Table~\ref{tab:split_generalization}. Its PE/CE SPL is 41.7/52.5 on seen
validation and 32.0/45.5 on unseen testing. The comparison uses different
scene sets and is not a paired change in the same environments.

\subsection{Goal Conditioned Tasks}
\label{app:goal-tasks}

\begin{table*}[h]
    \centering
    \caption{
        Generalization across instruction, object, and point goal tasks.
    }
    \label{tab:task_transfer}

    \vspace{3.5mm}

    \small
    \setlength{\tabcolsep}{7pt}
    \renewcommand{\arraystretch}{1.12}

    \begin{tabular}{@{}lcccc@{}}
        \toprule
        \multirow[c]{2}{*}{\textbf{Method / Task}}
        & \multicolumn{2}{c}{\textbf{Physical Environment}}
        & \multicolumn{2}{c}{\textbf{Continuous Environment}} \\
        \cmidrule(lr){2-3}
        \cmidrule(l){4-5}
        & SR$\uparrow$ & SPL$\uparrow$
        & SR$\uparrow$ & SPL$\uparrow$ \\
        \midrule

        Native VLN / Instruction
        & 26.8 & 17.3 & 41.8 & 29.9 \\
        Native VLN / Object
        & 20.0 & 12.0 & 28.1 & 18.4 \\
        Native VLN / Point
        & 48.3 & 35.8 & 63.0 & 50.7 \\

        \midrule

        BiLaLoRA / Instruction
        & 35.6 & 25.2 & 48.8 & 40.1 \\
        BiLaLoRA / Object
        & 25.3 & 16.1 & 34.3 & 23.5 \\
        BiLaLoRA / Point
        & 56.4 & 42.8 & 70.1 & 56.9 \\

        \midrule

        Restored RGB + Adapter / Instruction
        & 37.3 & 27.1 & 50.8 & 39.5 \\
        Restored RGB + Adapter / Object
        & 27.5 & 17.8 & 36.8 & 24.9 \\
        Restored RGB + Adapter / Point
        & 59.2 & 45.5 & 71.2 & 58.2 \\

        \midrule

        FireNav / Instruction
        & \textbf{41.6} & \textbf{32.0}
        & \textbf{56.1} & \textbf{45.5} \\
        FireNav / Object
        & \textbf{33.1} & \textbf{23.3}
        & \textbf{40.8} & \textbf{30.1} \\
        FireNav / Point
        & \textbf{63.0} & \textbf{50.4}
        & \textbf{74.5} & \textbf{63.6} \\

        \bottomrule
    \end{tabular}

    \vspace{-2mm}
\end{table*}

Table~\ref{tab:task_transfer} reports performance by task. Relative to
Restored RGB + Adapter, FireNav has 5.5/5.2 points higher PE/CE SPL on
object search and 4.9/5.4 points higher on point goal navigation. Without
holding out a task during training, these comparisons do not establish
transfer to unseen tasks.

\section{Visual History and Feature Reuse}
\label{app:history}

\begin{table*}[h]
    \centering
    \caption{
        Effect of visual history and feature caching.
    }
    \label{tab:history_cache}

    \vspace{3.5mm}

    \small
    \setlength{\tabcolsep}{4.6pt}
    \renewcommand{\arraystretch}{1.12}

    \begin{tabular}{@{}lcccccccc@{}}
        \toprule
        \multirow[c]{2}{*}{\textbf{Configuration}}
        & \multicolumn{2}{c}{\textbf{Cache}}
        & \multicolumn{2}{c}{\textbf{Runtime}}
        & \multicolumn{2}{c}{\textbf{PE}}
        & \multicolumn{2}{c}{\textbf{CE}} \\
        \cmidrule(lr){2-3}
        \cmidrule(lr){4-5}
        \cmidrule(lr){6-7}
        \cmidrule(l){8-9}
        & Calls
        & Hit (\%)
        & P50
        & P95
        & SR
        & SPL
        & SR
        & SPL \\
        \midrule

        No history, 3 current views
        & 3 & 0.0 & 206.1 & 256.5
        & 37.9 & 28.0 & 52.5 & 41.5 \\

        4 history + 3 current views
        & 3 & 57.1 & 242.1 & 302.2
        & 40.8 & 30.7 & 54.8 & 43.9 \\

        10 history + 3 current views
        & 3 & 76.9 & 277.9 & 348.9
        & 41.2 & 31.5 & 56.0 & 45.2 \\

        20 history + 3 current views
        & 3 & 87.0 & 310.1 & 389.4
        & \textbf{41.6} & \textbf{32.0}
        & \textbf{56.1} & \textbf{45.5} \\

        20 history, front-view evidence only
        & 1 & 95.2 & 292.1 & 369.5
        & 39.2 & 29.2 & 54.1 & 42.8 \\

        20 history, current evidence only
        & 3 & 0.0 & 306.5 & 384.8
        & 39.6 & 29.5 & 55.2 & 44.2 \\

        20 history, recompute all 23 images
        & 23 & 0.0 & 430.3 & 520.4
        & \textbf{41.6} & \textbf{32.0}
        & \textbf{56.1} & \textbf{45.5} \\

        \bottomrule
    \end{tabular}

    \vspace{-2mm}
\end{table*}

Moving from 10 to 20 history frames in Table~\ref{tab:history_cache} adds
0.5/0.3 PE/CE SPL points. With 20 frames, cached and recomputed features
have the same listed navigation scores, while P95 differs by 131.0 ms.
Restricting evidence to current views loses 2.5/1.3 SPL points.
Changing history length alters the available observations. Recomputing the
same features instead tests reuse, subject to identical inputs and update
scheduling.

\end{document}